\documentclass[sigconf]{acmart}

\copyrightyear{2025}
\acmYear{2025}
\setcopyright{acmlicensed}
\acmConference[MM '25] {Proceedings of the 33rd ACM International Conference on Multimedia}{October 27--31, 2025}{Dublin, Ireland.}
\acmBooktitle{Proceedings of the 33rd ACM International Conference on Multimedia (MM '25), October 27--31, 2025, Dublin, Ireland}
\acmISBN{979-8-4007-2035-2/2025/10}
\acmDOI{10.1145/XXXXXX.XXXXXX}

\AtBeginDocument{%
  }
    
\usepackage{multirow}
\usepackage{bm}
\usepackage{makecell}
\usepackage{algorithm}
\usepackage{algpseudocode}

\makeatletter
\newcommand*\bigcdot{\mathpalette\bigcdot@{.5}}
\newcommand*\bigcdot@[2]{\mathbin{\vcenter{\hbox{\scalebox{#2}{$\m@th#1\bullet$}}}}}
\makeatother   

\usepackage{booktabs}  
\usepackage{array}     
\begin{document}

\title{Accelerating Diffusion Transformer via Error-Optimized Cache}

\author{Junxiang Qiu}
\email{qiujx@mail.ustc.edu.cn}
\affiliation{%
  \institution{University of Science and Technology of China}
  \city{Hefei}
  \country{China}
}

\author{Shuo Wang}
\email{shuowang.edu@gmail.com}
\authornote{Correspond Author.}
\affiliation{%
  \institution{University of Science and Technology of China}
  \city{Hefei}
  \country{China}
}

\author{Jinda Lu}
\email{lujd@mail.ustc.edu.cn}
\affiliation{%
  \institution{University of Science and Technology of China}
  \city{Hefei}
  \country{China}
}

\author{Lin Liu}
\email{liulin0725@mail.ustc.edu.cn}
\affiliation{%
  \institution{University of Science and Technology of China}
  \city{Hefei}
  \country{China}
}

\author{Houcheng Jiang}
\email{jianghc@mail.ustc.edu.cn}
\affiliation{%
  \institution{University of Science and Technology of China}
  \city{Hefei}
  \country{China}
}

\author{Xingyu Zhu}
\email{xyzhuxyz@mail.ustc.edu.cn}
\affiliation{%
  \institution{University of Science and Technology of China}
  \city{Hefei}
  \country{China}
}

\author{Yanbin Hao}
\email{haoyanbin@hotmail.com}
\affiliation{%
  \institution{Hefei University of Technology}
  \city{Hefei}
  \country{China}
}

\renewcommand{\shortauthors}{Junxiang Qiu et al.}

\begin{abstract}

Diffusion Transformer (DiT) is a crucial method for content generation.
However, it needs a lot of time to sample. Many studies have attempted to use caching to reduce the time consumption of sampling.
Existing caching methods accelerate generation by reusing DiT features from the previous time step and skipping calculations in the next, but they tend to locate and cache low-error modules without focusing on reducing caching-induced errors, resulting in a sharp decline in generated content quality when increasing caching intensity.
To solve this problem, we propose the \textbf{E}rror-\textbf{O}ptimized \textbf{C}ache (\textbf{EOC}). This method introduces three key improvements: \textbf{(1)} Prior knowledge extraction: Extract and process the caching differences; \textbf{(2)} A judgment method for cache optimization: Determine whether certain caching steps need to be optimized; \textbf{(3)} Cache optimization: reduce caching errors. 
Experiments show that this algorithm significantly reduces the error accumulation caused by caching, especially excessive caching. On the ImageNet dataset, without substantially increasing the computational load, this method improves the FID↓ of the generated images when the rule-based model FORA has a caching level of \textbf{75}\%, \textbf{50}\%, and \textbf{25}\%, and the training-based model Learning-to-cache has a caching level of \textbf{22}\%. Specifically, the FID↓ values change from 30.454 to 21.690 (\textbf{28.8}\%), from 6.857 to 5.821 (\textbf{15.1}\%), from 3.870 to 3.692 (\textbf{4.6}\%), and from 3.539 to 3.451 (\textbf{2.5}\%) respectively. Code is available at \url{https://github.com/qiujx0520/EOC_MM2025.git}.
\end{abstract}





\maketitle

\section{Introduction}
\label{Introduction}


Content generation aims to automatically produce text \cite{achiam2023gpt, lu2025damo, zhu2024selective,li2025diffgad}, images \cite{xu2024imagereward, wang2025precise, lu2023semantic, zhu2024boosting, zhu2024enhancing}, videos \cite{wang2024lavie, qiu2025multimodal, wang2018connectionist, guo2019dense}, and other forms of content based on existing data or context. It is widely used in intelligent customer service dialogue generation, news summarization, and advertising copy creation. However, current generation methods often rely on complex models, resulting in cumbersome and time-consuming computational processes that hinder rapid deployment and flexible application in real-world business contexts. 


Recent approaches often employ strategies like pruning \cite{liu2018rethinking} or caching \cite{ma2024deepcache} to accelerate generative processes and tackle inference-speed and scalability challenges. Unlike pruning (removing unimportant parameters/neurons to simplify the model), caching methods can reuse intermediate computation results across multiple sampling or inference steps. This greatly boosts inference efficiency while preserving the model's expressive capacity. Common caching strategies fall into two broad types: rule-based \cite{selvaraju2024fora}, analyzing generated-content variations during sampling to decide what to cache/skip, and training-based \cite{ma2024learning}, enabling the model to automatically learn when, where, and how to cache/skip unimportant computational modules. These strategies are often integrated into the Diffusion Transformer (DiT) \cite{peebles2023scalable} model as the DiT has robust learning capabilities and maintains data dimensions during sampling, facilitating easy validation of these acceleration strategies. 




\begin{figure}[t]
\includegraphics[width=0.8\linewidth]{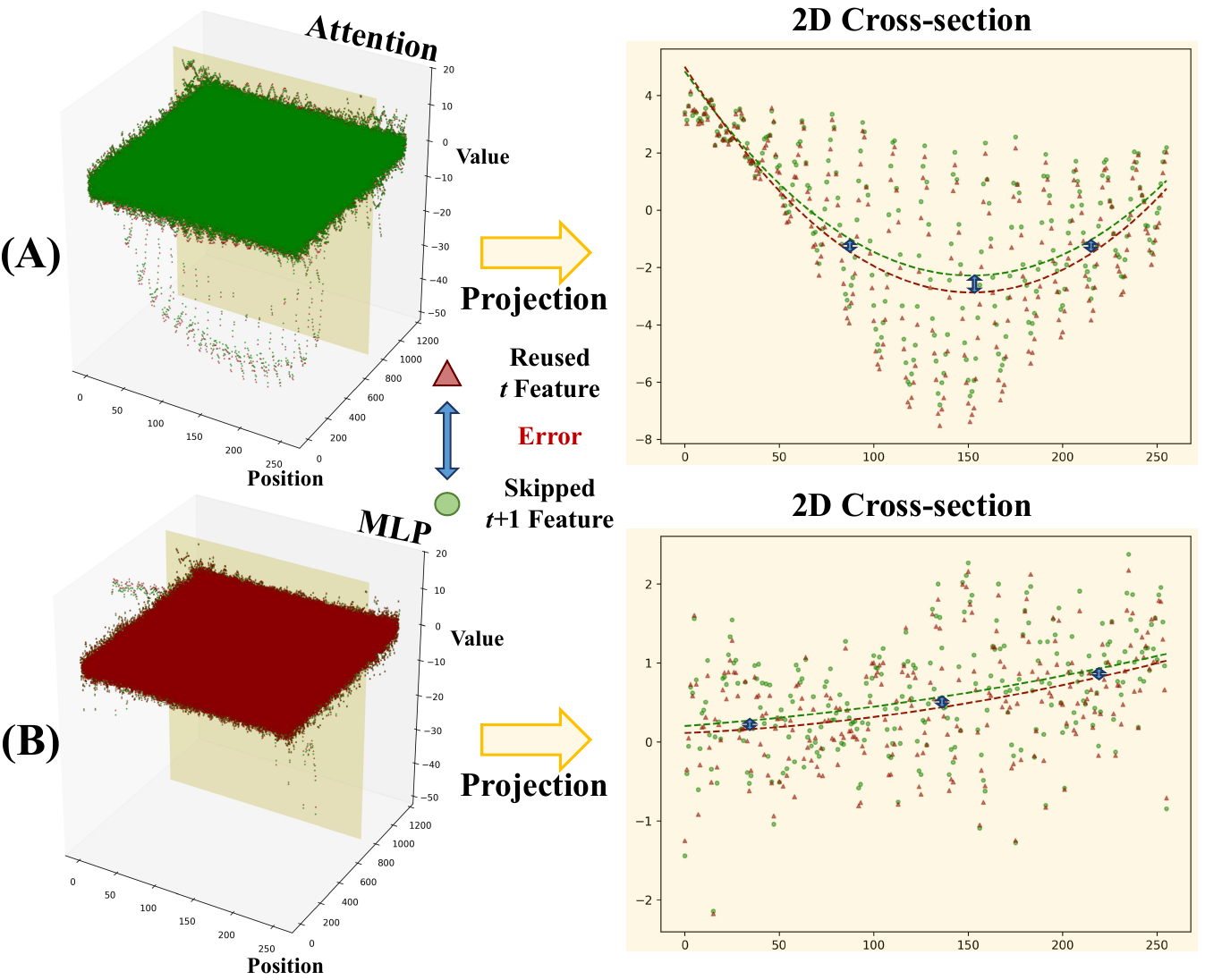}
\caption{The adjacent steps' outputs of (A) Attention and (B) MLP layers during the DiT calculation.
} 
\label{fig:block_view}
\end{figure}

\begin{figure}[t]
\includegraphics[width=0.7\linewidth]{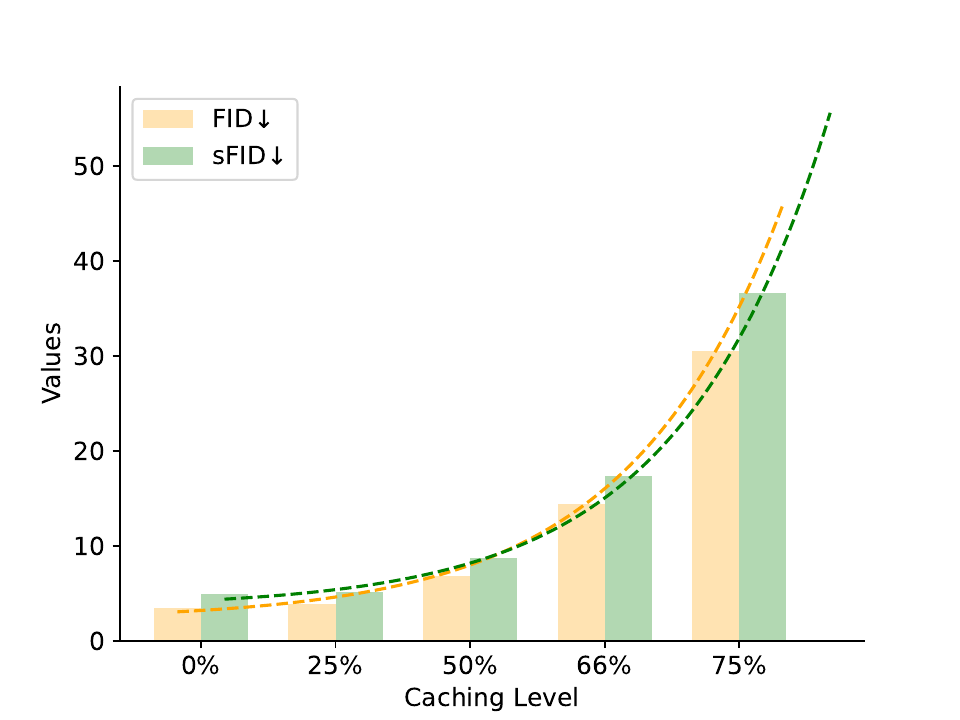}
\caption{Metrics of the content generated by DiT at different Caching levels (the percentage of reused blocks).} 
\label{fig:ehua}
\end{figure}

However, the caching-based DiT method typically reuses the same computational blocks across consecutive steps without sufficiently analyzing potential errors in the cached content, leading to cumulative errors at each caching operation. To verify this, we visualize the outputs of the Attention and MLP layers from the DiT computation process in Figures~\ref{fig:block_view}(A) and~\ref{fig:block_view}(B), focusing on the same blocks in adjacent steps. Specifically, the red dots represent the previous step, while the green dots correspond to the current step. Although most points from these two steps lie on a shared hyperplane, numerous outliers exhibit substantial deviations. We project one layer into a 3D structural diagram to further illustrate these discrepancies. We can find that even within that hyperplane, there are discernible gaps between the two point sets. Thus, in conventional DiT methods, the caching process typically begins by statistically analyzing how caching specific modules affects generation quality, aiming to identify those with minimal impact. The method then replaces the current step’s outputs (green) with those from the previous step (red) to reduce computation. However, this replacement process disregards the underlying differences, amplifying errors through repeated reuse. Additionally, the absence of thorough error analysis requires careful selection of which modules to cache. Over-caching can severely degrade both the quality and stability of the generated content. As shown in Figure~\ref{fig:ehua}, we further validated these observations using a rule-based DiT method \cite{selvaraju2024fora} and tested caching at 0\%, 25\%, and 50\%. While caching 25\% of the blocks yields results comparable to those without caching, caching 50\% of the blocks noticeably degrades generation quality. This decline occurs because blocks with large errors, once cached, accumulate and manifest as noise or extraneous artifacts, ultimately compromising the overall quality of the generated content.

Inspired by the concept of knowledge editing \cite{meng2022locating}, we propose introducing different values from corresponding positions in prior knowledge to mitigate significant caching errors, thereby reducing the negative impacts of caching. Prior knowledge refers to the outputs of the attention layer and the MLP layer of each DiT block at every step of the sampling process. Specifically, we begin by determining which caching modules require optimization, measuring two factors: (1) the magnitude of error introduced when caching a given module, and (2) the relative position of that module's step in the overall sampling process. Larger errors, especially if they occur early in sampling, are more prone to accumulation. By weighting and summing these two factors, we derive an indicator to identify modules that demand caching optimization. Once such modules are identified, we introduce controlled perturbations into the cached content during sampling to further reduce errors. Specifically, when a module is selected for caching, we calculate the ``trend'' at the same position in the prior knowledge and embed it into the caching operation via weighted multiplication. This effectively maps the prior knowledge’s trend to the sampled data, helping to maintain the stability and quality of the generated results while benefiting from the computational efficiency of caching.
In summary, the contributions of Error-Optimized Cache (EOC) are threefolds:
\begin{itemize}
\item We design a module that extracts and processes the model’s operational data, capturing the characteristics of changes between blocks. It can be reused in different caching methods under the same model.
\item We propose a method that determines when caching steps require optimization. This ensures sufficient optimization and prevents over-optimization of caching from causing blurring of details.
\item Our approach reduces caching errors, enhances the robustness of cached models and raises the upper bound on image generation speed without sacrificing quality. 
\end{itemize}

\section{Related Work}
\label{Related WorK}

In this section, we review existing diffusion acceleration methods and outline the differences between our proposed EOC approach and these related techniques.

\subsection{Traditional Generation Acceleration}
Common methods for sampling acceleration include pruning, quantization, and sampling method optimization.

\textbf{Model Pruning And Quantization}. {Pruning} reduces model complexity and computational cost by removing less critical weights, neurons, or layers while maintaining performance. It involves unstructured pruning \cite{dong2017learning, lee2019signal}, which masks individual parameters, and structural pruning \cite{liu2021group}, which removes larger components like layers or filters. In diffusion models, techniques like LD-Pruner \cite{castells2024ld} and DiffPruning \cite{guo2020parameter} enhance efficiency by applying structured pruning and using informative gradients to reduce unnecessary steps.
{Quantization} lowers the precision of weights and activations to smaller bit formats, reducing model size and speeding up inference. Key methods include Quantization-Aware Training (QAT) \cite{bhalgat2020lsq+} and the more efficient Post-Training Quantization (PTQ) \cite{li2021brecq, nagel2020up}, which doesn’t require retraining. For diffusion models, PTQ methods such as PTQ4DiT \cite{wu2024ptq4dit} and Q-diffusion \cite{li2023q} optimize activation variance across denoising steps, while TDQ \cite{so2024temporal} uses MLP layers for step-wise parameter estimation. MPQ-DM\cite{feng2025mpq} achieves significant accuracy gains at extremely low bit-widths through two techniques: outlier-driven mixed quantization (OMQ) and temporal smooth relationship distillation (TRD). These approaches aim to balance performance with efficiency.

\textbf{Sampling Optimization}. Sampling method optimization for diffusion models aims to reduce generation time while maintaining sample quality. These can be categorized into re-training methods, like knowledge distillation \cite{luhman2021knowledge, salimans2022progressive, feng2024relational}, and those that do not require re-training, such as advanced samplers for pre-trained models. Notably, DDIM \cite{song2020denoising} reduces sampling steps using non-Markovian processes, while DPM-Solver \cite{lu2022dpm} leverages ODE solvers. Progressive distillation \cite{meng2023distillation, yin2024one}, consistency models \cite{song2023consistency}, and parallel sampling methods like DSNO \cite{zheng2023fast} and ParaDiGMS \cite{shih2024parallel} also enhance efficiency without additional training, making diffusion models more practical for real-world applications.

\begin{figure*}[t]
\includegraphics[width=1\textwidth]{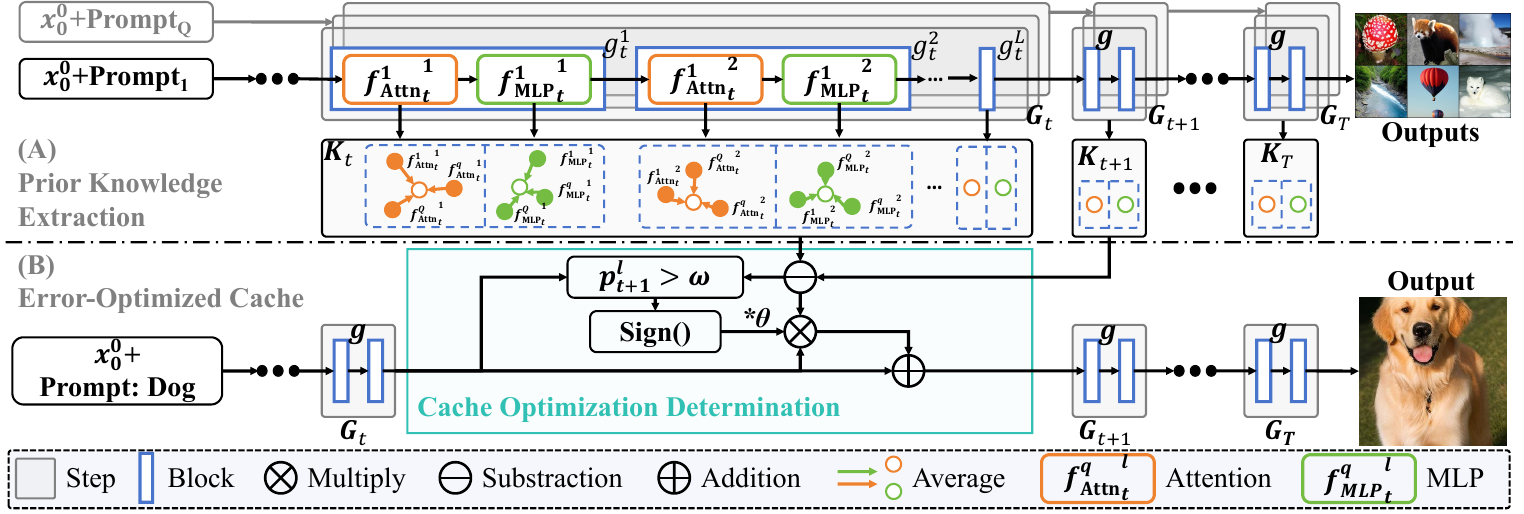}
\caption{Pipeline of EOC. (A) Prior Knowledge Extraction: Multiple inputs with different prompts are fed into the sampling process, and the outputs $\bm{f}$ of the Attention layers and MLP layers are recorded;
(B) Cache Optimization: By examining the difference between the knowledge $\bm{K}_{t}$ and $\bm{K}_{t+1}$ of adjacent steps, as well as the step position, it is determined whether to perform Cache Optimization at the current step. 
} 
\label{fig:pipeline}
\end{figure*}

\subsection{Caching Method}
In addition to the methods above, caching offers another low-cost and highly adaptable approach to accelerate generation. Initially, Faster Diffusion \cite{li2023faster} and DeepCache \cite{ma2024deepcache} observed the sampling process to cache and reuse module outputs with minimal loss. Cache-Me-if-You-Can \cite{wimbauer2024cache} uses a teacher-student model to mimic the original model's sampling process, reducing caching errors. However, these techniques are designed for U-Net-based models \cite{song2020denoising} and are difficult to apply directly to DiT-based models \cite{peebles2023scalable}.

Next, Fora \cite{selvaraju2024fora} implemented a caching mechanism that stores and reuses intermediate outputs of attention and MLP layers across denoising steps, reducing computational overhead. $\Delta$-DiT \cite{chen2024delta} uses a designed caching mechanism to accelerate the later DiT blocks in the early sampling stages and the earlier DiT blocks in the later stages. These works extend caching to DiT-based models. ToCa \cite{zou2024accelerating} extends the caching dimension from the block level to the token level, thereby increasing the upper limit of caching acceleration. Beyond the rule-based methods mentioned, learning-to-cache \cite{ma2024learning} achieves higher acceleration by using a learnable router to decide whether each layer needs computation, though it incurs significant computational costs. Building on image generation, caching can also be applied to video generation acceleration\cite{zhao2024real, lv2024fastercache, liu2025timestep}.

\noindent \textbf{Differences:} The above-mentioned methods place more emphasis on using different approaches to locate the unimportant (cacheable) parts of the model, thereby achieving a balance between generation acceleration and generation quality. However, they lack consideration for the bias of the cached content. In contrast, \textbf{(1)} the EOC method proposed in this paper can maximally reduce the problem of error accumulation caused by caching errors and can be applied to both rule-based and training-based methods.
\textbf{(2)} The above-mentioned methods have not explored extracting the prior knowledge of the model and applying it in the sample process.


\section{Method}
\label{Method}
In this section, we first introduce our work by presenting the DiT and Cache methods. Subsequently, we elaborate on the three main steps encompassed in EOC: \textbf{(1)} Prior Knowledge Extraction,
\textbf{(2)} Cache Optimization Determination,
and \textbf{(3)} Cache Optimization.

\subsection{Preliminaries}
\textbf{Diffusion Models}. Diffusion models \cite{ho2020denoising} contain: (1) The forward process: By gradually adding Gaussian noise, the original image $\bm{x}_0$ is added with noise to become the random noise $\bm{x}_T$, where $T$ represents the total number of denoising steps; (2) The reverse process: the Gaussian random noise $\bm{x}_T$ is gradually denoised to reconstruct the original image $\bm{x}_0$. To achieve the step-by-step process, the model needs to learn the inverse process of noise addition. Therefore, we can model the reverse process $\mathcal{R}(\bm{x}_{t-1} \mid \bm{x}_{t})$ using Markov chains \cite{norris1998markov} $\bm{\mathcal{N}}$ as follows: 
\begin{equation}
\mathcal{R}(\bm{x}_{t-1} \mid \bm{x}_{t}) = \bm{\mathcal{N}} \left( \bm{x}_{t-1}; \frac{1}{\sqrt{\alpha_t}} \left( \bm{x}_t-\frac{1-\alpha_t}{\sqrt{1-\bar{\alpha}_t}} \bm{\epsilon}(\bm{x}_t, t) \right) \right),
\label{eq:1}
\end{equation}
where $t$ represents the denoising step, $\alpha_t$ represents the noise variance schedule, and $\bar{\alpha}_t = \prod_{i=1}^{T} \alpha_i$.
$\bm{\epsilon}$ is a designed deep network. It takes $\bm{x}_t$ as input and outputs a prediction of the noise required for the denoising process. During the reverse process with $T$ timesteps, $\bm{\epsilon}$ is employed for inference $T$ times. 

\textbf{Diffusion Transformer}. Diffusion Transformer (DiT) \cite{peebles2023scalable}, typically consists of stacked groups of attention layer and multilayer perceptron. 
Mathematically, it can be approximately described by the following formulation:
\begin{equation}
\begin{split}
\bm{G}_t&=\bm{g}_t^1\circ \bm{g}_t^2\circ\cdots\circ \bm{g}_t^l\circ\cdots\circ\bm{g}_t^L,\quad\text{where}\\
\bm{g}_t^l&={\bm{F}_{\mathrm{Attn}}}_t^l\circ{\bm{F}_{\mathrm{MLP}}}_t^l,
\end{split}
\label{eq:2}
\end{equation}
where $\bm{G}_t$ represents timestep $t$ in the DiT model, $\bm{g}_t^l$ stands for the individual DiT block, $\circ$ represents data transfer and $({\bm{F}_{\mathrm{Attn}}}_t^l, {\bm{F}_\mathrm{MLP}}_t^l)$ denotes the different modules in a single DiT block. The computational output dimensions of these modules and layers are all the same. The variable $l$ serves as the index for the DiT blocks, while $L$ represents the depth of the DiT model.
Before delving into the general computation between $({\bm{F}_{\mathrm{Attn}}}_t^l, {\bm{F}_\mathrm{MLP}}_t^l)$ and $\bm{x}_t^l$, it's essential to introduce ALN \cite{guo2022adaln}, referring to the adaptive layer normalization. It has become an important component in modern neural network architectures, especially in the context of DiT.
Subsequently, the general computation between them can be written as:
\begin{equation}
\begin{split}
{\bm{F}_{\mathrm{Attn}}}_t^l&=\bm{x}_t^l+\text{ALN}\circ {\bm{f}_{\mathrm{Attn}}}_t^l(\bm{x}_t^l),\\
\bm{x}_{t+1}^l={\bm{F}_{\mathrm{MLP}}}_t^l&={\bm{F}_\mathrm{Attn}}_t^l+\text{ALN}\circ {\bm{f}_{\mathrm{MLP}}}_t^l({\bm{F}_{\mathrm{Attn}}}_t^l),
\end{split}
\label{eq:f}
\end{equation}
 where, $\bm{x}_t^l$ and ${\bm{F}_{\mathrm{Attn}}}_t^l$ represent the residual connection. And the output of attention and MLP is $({\bm{f}_{\mathrm{Attn}}}_t^l(\bm{x}_t), {\bm{f}_\mathrm{MLP}}_t^l({\bm{F}_{\mathrm{Attn}}}_t^l))$. And ALN is applied to the output. By normalizing these values adaptively, ALN helps to stabilize the training process, improve the model's generalization ability, and enhance the overall performance of the DiT model in handling complex data patterns.

\textbf{Feature Caching}. 
Feature caching is centered on the principle of re-utilizing features that were calculated and stored during earlier timesteps. The primary objective is to avoid redundant computations in the subsequent timesteps. This caching mechanism generally operates through multiple, repeated caching intervals.
Let's assume there is a caching interval that encompasses $N$ timesteps, starting from timestep $t$ and ending at timestep $t + N-1$. In this scenario, in timestep $t$, feature caching performs a full-fledged calculation and then stores the generated features. Symbolically, we can represent this process as $\bm{C}[l]:=\bm{g}_t^l(\bm{x}_t^l)$. Here, $\bm{C}[l]$ denotes the cache, $\bm{g}_t^l(\bm{x}_t^l)$ corresponds to $({\bm{f}_{\mathrm{Attn}}}_t^l(\bm{x}_t), {\bm{f}_\mathrm{MLP}}_t^l({\bm{F}_{\mathrm{Attn}}}_t^l))$ of the $l$-th block at timestep $t$, and the symbol “$:=$” represents an assignment operation.
After that, for the timesteps $[t + 1, t + N-1]$, instead of carrying out new computations, the system leverages the pre-stored features. Mathematically, this can be formulated as $\bm{g}_t^l(\bm{x}_{t + i}^l):=\bm{C}[l]$, where $i$ is a non-negative integer and $i\leq{N-1}$.

\subsection{Error-Optimized Cache}
For feature cache, previous work has centered on using different methods to locate modules that are either irrelevant or weakly related to sampling for caching, thus accelerating the generation process. However, they lack an exploration of caching errors. As a result, all these methods have a relatively low acceleration ceiling. Therefore, our goal is to introduce low-cost prior knowledge with virtually no time cost, to offset the relatively large cache error to the greatest extent, and to optimize the generation quality at the same cache intensity. As shown in Figure~\ref{fig:pipeline} (A), to achieve this goal, first, we need to obtain the targeted output trends of each block through pre-sampling. And as shown in Figure~\ref{fig:pipeline} (B), we need to determine whether a module to be cached requires optimization. Finally, we also need to embed this prior information into the sampling process. Next, we will introduce these three methods in detail.

\textbf{Prior Knowledge Extraction}.
\label{PKE}
In order to determine whether to perform cache optimization, we need to record every outputs during the sampling process.
Specifically, we will use a subset of prompts that cover a portion of the categories. These will be input into the model separately and multiple times for original generation without caching. As shown in Figure~\ref{fig:pipeline} (A), during each generation process, we add hooks to record the outputs $({\bm{f}_{\mathrm{Attn}}}_t^l(\bm{x}_t), {\bm{f}_\mathrm{MLP}}_t^l({\bm{F}_{\mathrm{Attn}}}_t^l))$. Subsequently, we will calculate the average of the Attention and MLP outputs at each position respectively, so as to obtain representative features ${\bm{K}_{\mathrm{Attn}}}_t^l$ and ${\bm{K}_{\mathrm{MLP}}}_t^l$ of the model generation process. Assuming that the sampling is performed $Q$ times. The specific calculation functions are shown as follows:
\begin{equation}
\begin{split}
{\bm{K}_{\mathrm{Attn}}}_t^l = \frac{1}{Q} \sum_{q = 1}^{Q} {\bm{f}_{\mathrm{Attn}}^{q}}_t^l, \ \ 
{\bm{K}_{\mathrm{MLP}}}_t^l = \frac{1}{Q} \sum_{q = 1}^{Q} {\bm{f}_{\mathrm{MLP}}^{q}}_t^l.
\end{split}
\end{equation}

\textbf{Cache Optimization Determination}.
To determine whether a block to be cached requires cache optimization, we believe that two points need to be focused on: 
(1) If caching a certain block has a significant negative impact on the model, or in other words, if the output of the block whose calculation is to be skipped differs greatly from the output of the block whose cache is to be reused, then we will prioritize the optimization of these caches. 
(2) If cache optimization is performed on a block, and the embedded optimized content will lead to wrong details or blurred background in the model output, then we will avoid optimizing these caches.

First, we analyze the prior knowledge extracted in Section~\ref{PKE}. In the previous timestep $t$ and subsequent timesteps $t+i$, the blocks at the same position can be regarded as the modules to be calculated, cached, and reused, and the modules to be skipped, respectively. Therefore, by calculating the differences of the corresponding modules, we can obtain the changing trend of each pixel point, which is collectively referred to as ``trend'', namely ${\bm{E}_{\mathrm{Attn}}}_t^l$ and ${\bm{E}_{\mathrm{MLP}}}_t^l$. The specific calculation functions are shown as follows:
\begin{equation}
{\bm{E}_{\mathrm{Attn}}}_{t+1}^l = {\bm{K}_{\mathrm{Attn}}}_{t+1}^l - {\bm{K}_{\mathrm{Attn}}}_t^l, {\bm{E}_{\mathrm{MLP}}}_{t+1}^l = {\bm{K}_{\mathrm{MLP}}}_{t+1}^l - {\bm{K}_{\mathrm{MLP}}}_t^l, 
\label{eq:ftoa}
\end{equation}
for the ``trends'' ${\bm{E}_{\mathrm{Attn}}}_t^l$ and ${\bm{E}_{\mathrm{MLP}}}_t^l$, we calculate the average value ${v}_t^l$ of all pixel points after taking their absolute values:
\begin{equation}
{v}_t^l = \langle \bm{J},|{\bm{E}_{\mathrm{Attn}}}_{t+1}^l| + |{\bm{E}_{\mathrm{MLP}}}_{t+1}^l|\rangle_\mathcal{F}/{\textstyle\sum \bm{J}},
\end{equation}
where $\bm{J}$ is an all-one matrix of the same size as $\bm{E}$, $|\bigcdot|$ is the absolute value operation, $\langle\bigcdot,\bigcdot\rangle_\mathcal{F}$ is Frobenius Inner Production \cite{montero2002approximate}.
Large average value ${v}_t^l$ indicates a large error between the corresponding blocks. We will apply cache optimization for the modules with larger ${v}_t^l$.

\begin{figure*}[t]
    \includegraphics[width=0.9\textwidth]{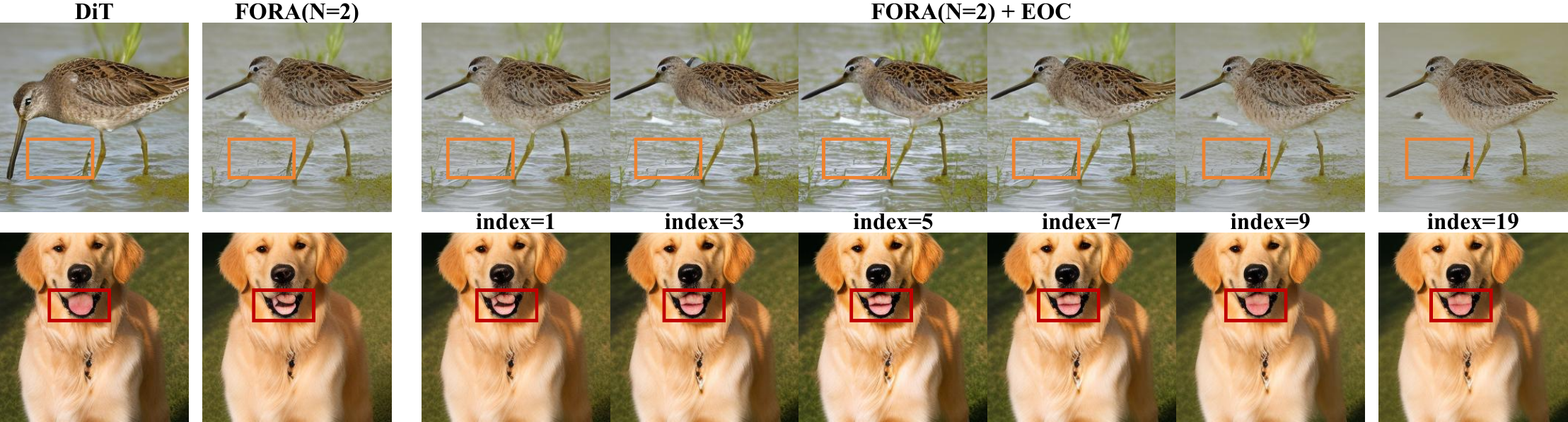}
    \caption{The images of birds and dogs generated after performing the CO operation on the cached blocks where \(t\leq{index}\).} 
    \label{fig: xijie+beijing}
\end{figure*}

Next, through observation, the absolute values of each pixel value in the trend vary. For relatively large ${v}_t^l$, we believe that they are the accumulation of many numbers with the same sign (positive or negative). Therefore, large numbers are representative of the trend and can be used to compensate for the errors caused by caching and reduce the accumulation of cache errors. As shown in Figure~\ref{fig: xijie+beijing}, the $index$ represents that cache optimization is applied in the first $index$ steps. The dog's tongue is repaired as the index increases. 
For relatively small ${v}_t^l$, we consider it to be an accumulation of positive and negative values around zero. If they are applied in the later stage of generation, they will cause problems such as blurred background details. As shown in Figure~\ref{fig: xijie+beijing}, when $index=19$, the details of the water in the bird's background and the grass in the dog's background are greatly weakened. 
In other words, even if trends are introduced in the early stage of the sampling process, causing errors, the model can map these errors back to the normal range through subsequent sampling steps. However, errors caused by introducing trends in the later stage do not have enough time to return to normal. Therefore, introducing trends in the later stage will result in more negative impacts.
Therefore, the step at which cache optimization is applied should be in the early stage of generation. In this way, through model operations, the noise with small absolute values will be mapped into the required generated content, thus avoiding the above-mentioned problems. 

Based on the above analysis, we take the magnitude of the error ${v}_t^l$ and the step position ${t}/{T}$ of caching as important judgment conditions for cache optimization determination. In response to this, we design the following formula to rank the cache optimization priorities ${p}_t^l$ of different blocks:
\begin{equation}
{p}_t^l = 
\begin{cases} 
\gamma \cdot {v}_t^l + (1-\gamma) \cdot (1-t/T), & Cache, \\
0, & No\ Cache,
\end{cases}
\label{eq:3}
\end{equation}
where $\gamma$ represents a regulation coefficient, which is used to adjust the importance of the error and step positions. And we don't distinguish between the Attention layers and the MLP layers, as they both contribute to the calculation of ${v}_t^l$. We set a threshold $\omega$, which is used to adjust cache optimization level. When ${p}_t^l > \omega$, we perform cache optimization on the $l$-th block at the timestep $t$.

\textbf{Cache Optimization}.
After completing the cache optimization determination and deciding that a certain block requires cache optimization, we need to consider how to appropriately embed the ``trend'' (${\bm{E}_{\mathrm{Attn}}}_t^l$ and ${\bm{E}_{\mathrm{MLP}}}_t^l$) from the prior knowledge into the sampling process to reduce cache errors. Since the original intention of caching is to reduce the time consumption of sampling, embedding the ``trend'' can only involve a minimal amount of computation. Additionally, the ``trend'' only contains the common information of a certain model's generation, while the generation also depends on the input class or prompt. Therefore, it is necessary to map the ``trend'' to the features in the generation process, fuse them together, and then offset the errors caused by caching.

Regarding the issue of time consumption, the prior knowledge used in EOC is extracted in advance and does not take up the time of the sampling process. Moreover, we do not pre-process the ``trend'' using the process information from sampling. As for embedding the ``trend'', after weighting it, we first multiply it by the cache information at the corresponding position and then add it to the sampling process. Eq.~\eqref{eq:f} can be written in the following form:
\begin{equation}
\begin{split}
{\bm{F}_{\mathrm{Attn}}}_t^l &= \bm{x}_{t}^l + \text{ALN} \circ (\bm{C}_\mathrm{Attn}[l] \cdot (1 + \theta \cdot {\bm{E}_{\mathrm{Attn}}}_t^l)), \\
\bm{x}_{t+1}^l={\bm{F}_{\mathrm{MLP}}}_t^l &= {\bm{F}_{\mathrm{Attn}}}_t^l + \text{ALN} \circ (\bm{C}_\mathrm{MLP}[l] \cdot (1 + \theta \cdot {\bm{E}_{\mathrm{MLP}}}_t^l)),
\end{split}
\label{eq:4}
\end{equation}
where $\theta$ represents the parameter for adjusting the magnitude of the ``trend''. 
We also provide Algorithm~\ref{algorithm: 1} to better describe our EOC method.

\begin{algorithm}[t]
\caption{EOC Block Algorithm}
\begin{algorithmic}[1]
\Require $\bm{x}_{t}^l$, $\bm{E}_{t}^l$, $\bm{C}[l]$, $\theta$, ${\bm{f}_\mathrm{Attn}}_{t}^l(\bigcdot)$, ${\bm{f}_\mathrm{MLP}}_{t}^l(\bigcdot)$, $t$
\Ensure $\bm{x}_{t+1}^l$, $\bm{C}[l]$
\If{No Cache}
    \State same as Eq.~\eqref{eq:f}
\Else{ Cache}
    \If{ No EOC}
        \State $\bm{C}_{\mathrm{Attn}}[l], \bm{C}_{\mathrm{MLP}}[l] \leftarrow \bm{C}[l]$
        \State ${\bm{F}_{\mathrm{Attn}}}_t^l \leftarrow \bm{x}_{t}^l + \text{ALN} \circ \bm{C}_{\mathrm{Attn}}[l]$
        \State $\bm{x}_{t+1}^l \leftarrow {\bm{F}_{\mathrm{MLP}}}_t^l \leftarrow {\bm{F}_{\mathrm{Attn}}}_t^l + \text{ALN} \circ \bm{C}_{\mathrm{MLP}}[l]$
    \Else{ EOC}
        \State same as Eq.~\eqref{eq:4}
    \EndIf
\EndIf
\State \Return $\bm{x}_{t+1}^l, \bm{C}[l] \leftarrow ({\bm{f}_{\mathrm{Attn}}}_t^l(\bm{x}_t^l), {\bm{f}_\mathrm{MLP}}_t^l({\bm{F}_{\mathrm{Attn}}}_t^l))$
\end{algorithmic}
\label{algorithm: 1}
\end{algorithm}

\begin{figure*}[t]
\includegraphics[width=0.8\textwidth]{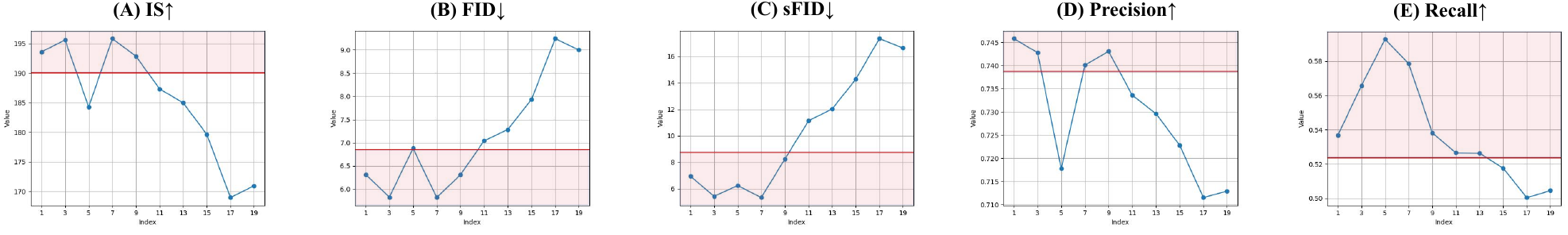}
\caption{The generation quality (A)-(E) after performing cache optimization on cached blocks with $t\leq{index}$.} \label{fig: fullcache}
\end{figure*}

\section{Experiment}
\label{Experiment}
In this section, we evaluate the proposed EOC method and compare it with rule-based and training-based methods. We also use ablation experiments to verify the effectiveness of the method proposed in this paper.\\
\textbf{RQ1}: How to determine if a block needs cache optimization?\\
\textbf{RQ2}: What kind of cache optimization should be performed?\\
\textbf{RQ3}: How does EOC compare with other methods?


\begin{table*}[t]
    \centering
    \caption{Comparison of cache optimization positions. (1) Apply the optimization to the Attention layers and the MLP layers respectively; (2) Apply the optimization to both the Attention layers and the MLP layers simultaneously.}\label{tab3.1}
    \begin{tabular}{c|c|c|ccccc}
        \toprule
        \textbf{Method} & \textbf{Caching level} & \textbf{$m$ Type} & \textbf{IS↑} & \textbf{FID↓} & \textbf{sFID↓} & \textbf{Precision↑} & \textbf{Recall↑} \\ 
        \midrule        
        \multirow{4}{*}{FORA + EOC} & \multirow{4}{*}{25\%} & \textbf{N/A} & \textbf{220.011} & 3.870 & 5.185 & \textbf{0.783} & 0.569 \\ 
        & & \textbf{Attn+MLP} & 218.363 & \textbf{3.692} & \textbf{5.122} & 0.781 & 0.585 \\ 
        & & \textbf{MLP} & 207.959 & 4.373 & 6.019 & 0.758 & \textbf{0.601} \\ 
        & & \textbf{Attn} & 207.425 & 5.597 & 11.148 & 0.764 & 0.519 \\ 
        \midrule
        \multirow{4}{*}{L2C + EOC} & \multirow{4}{*}{22\%} & \textbf{N/A} & \textbf{225.004} & 3.539 & 4.710 & 0.788 & 0.563 \\ 
        & & \textbf{Attn+MLP} & 223.957 & \textbf{3.451} & \textbf{4.675} & 0.789 & 0.570 \\ 
        & & \textbf{MLP} & 218.314 & 3.725 & 5.404 & 0.777 & \textbf{0.588} \\ 
        & & \textbf{Attn} & 224.705 & 3.909 & 5.645 & \textbf{0.792} & 0.543 \\ \bottomrule
    \end{tabular}
\end{table*}

\subsection{Experiment Settings}

\noindent\textbf{Datasets}. In this experiment, we selected the ImageNet \cite{deng2009imagenet} (1k classes)  dataset and the MS-COCO \cite{lin2014microsoft} (30k text prompts) dataset for testing. Unless otherwise specified that the MS-COCO dataset is used, the ImageNet dataset is used by default. 

\noindent\textbf{Model Configuration}. For DiT (class-to-image), we default to using DDIM \cite{song2020denoising} for 20-steps or 50-steps sampling in all experiments. For Pixart-$\alpha$ \cite{chen2023pixart} (text-to-image), we default to using DPM-Solver \cite{lu2022dpm} for 20-steps sampling in all experiments. In terms of generated image quality evaluation, we employed the IS, FID, sFID, Precision, and Recall metrics. The computational cost of the inference process is represented by FLOPs. When comparing different caching and cache optimization scenarios, we randomly generated 50k or 30k samples for ImageNet and MS-COCO evaluation, respectively. During the experiment, we utilized an A40 GPU and an Intel(R) Xeon(R) Gold 6248R CPU to meet the computational requirements.

\begin{table}[t]
\centering
\caption{Comparison of the effects of using addition and multiplication to embed prior knowledge when 25\% of the blocks are cached. }\label{tab3.2}
\begin{tabular}{c|ccc}
\toprule
\textbf{CO Strategy}                                                 & \textbf{IS↑}                & \textbf{FID↓}              & \textbf{sFID↓}             \\ \midrule
No Cache                                                      & 223.490                      & 3.484                      & 4.892                      \\ \midrule
FORA \cite{selvaraju2024fora}                                               & \textbf{220.011} & 3.870         & 5.185          \\  
FORA-``×'' & 218.363           & \textbf{3.692} & \textbf{5.122} \\ 
FORA-``+'' & 218.809           & 3.900         & 5.181          \\ \bottomrule
\end{tabular}
\end{table}

\begin{figure*}[t]
\includegraphics[width=0.8\textwidth]{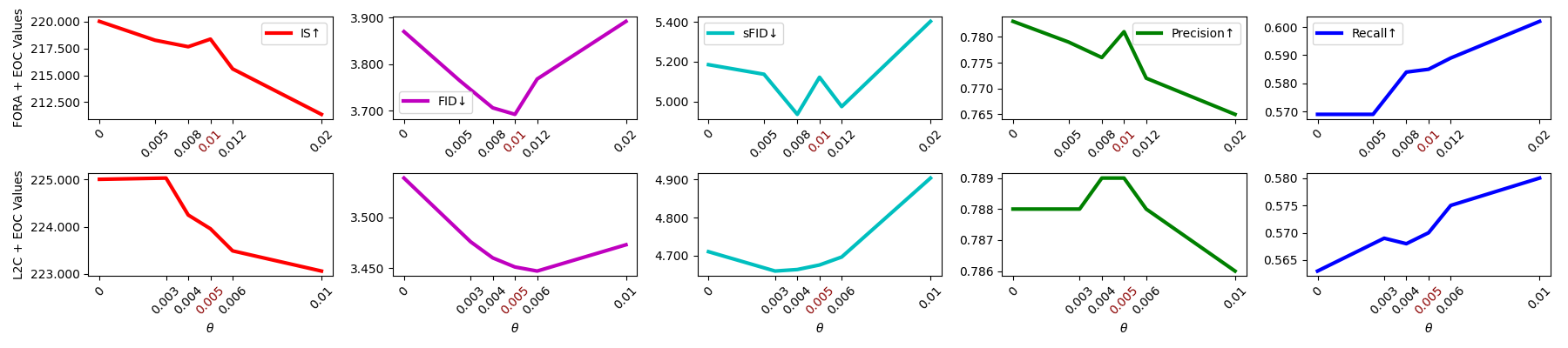}
\caption{Comparing the generation quality obtained with different values of $\theta$ under FORA and L2C.} 
\label{fig: theta}
\end{figure*}

We introduce EOC to verify its effectiveness based on the work of FORA \cite{selvaraju2024fora}, Learning-to-cache (L2C) \cite{ma2024learning}, ToCa \cite{zou2024accelerating}, and others. The baselines include Rule-based Method and Training-based Method. For the convenience of experimental presentation, we also provide the abbreviations for Cache Optimization Determination (COD) and Cache Optimization (CO). Unless otherwise specified, the caching period is defaulted to $ N = 2 $.

\begin{table*}[t]
\centering
\caption{Comparing the generation quality of FORA and L2C before and after applying EOC. And FORA includes two scenarios: normal-intensity caching (25\%) and over-caching (50\%).}\label{tab4.1}
\begin{tabular}{c|c|ccccc|cc}
\toprule
\textbf{Method} & \textbf{Caching level} & \textbf{IS↑} & \textbf{FID↓} & \textbf{sFID↓} & \textbf{Precision↑} & \textbf{Recall↑} & \textbf{FLOPs(T)↓} & \textbf{Speed↑} \\ \midrule
DiT \cite{peebles2023scalable} & 0 & 223.490 & 3.484 & 4.892 & 0.788 & 0.571 &11.868 &1× \\ \midrule
\multirow{2}{*}{FORA \cite{selvaraju2024fora}} & 25\% & \textbf{220.011} & 3.870 & 5.185 & \textbf{0.783} & 0.569 &8.900 &1.3335× \\  
& 50\% & 190.046 & 6.857 & 8.757 & 0.739 & 0.524 &5.934 &2.0000× \\ \midrule
\multirow{2}{*}{FORA + EOC} & 25\% & 218.363 & \textbf{3.692} & \textbf{5.122} & 0.781 & \textbf{0.585} &8.901 &1.3333× \\  
& 50\% & \textbf{195.844} & \textbf{5.821} & \textbf{5.342} & \textbf{0.740} & 0.579 &5.935 &1.9997× \\ \midrule
L2C \cite{ma2024learning} & 22\% & \textbf{225.004} & 3.539 & 4.710 & 0.788 & 0.563 &9.257 &1.2821× \\ \midrule
L2C + EOC & 22\% & 223.957 & \textbf{3.451} & \textbf{4.675} & \textbf{0.789} & \textbf{0.570} &9.258 &1.2819× \\ \bottomrule
\end{tabular}
\end{table*}

\subsection{Cache Optimization Determination}
\textbf{Effect of COD (RQ1).} Regarding cache-optimization determination of which step positions to prioritize for cache optimization, we design an experiment as shown in Figure~\ref{fig: fullcache}. This figure has five sub-graphs (A)-(E), representing IS, FID, sFID, Precision, and Recall respectively. To establish a base model, we apply FORA(N=2) caching strategy to DiT. Specifically, all odd-numbered steps with $t\in\{1, 3,..., 19\}$ were cached. The value of this base model is represented by the red horizontal line in each graph. The abscissa index of the ten blue dots in the figure represents cache optimization is performed on steps where $t\leq{index}$. 

Through observation, we can find that if we perform cache optimization when $t\leq19$, cache optimization will have a significant negative impact on the model's generation. As the value of $index$ gradually decreases, the model's generation performance gradually improves. When $index\in\{7, 9\}$, the model outperforms the base model in every evaluation criterion, and the effect of cache optimization is the best at this time. Therefore, we should try our best to avoid performing cache optimization on blocks with later steps.

\subsection{Cache Optimization}
\textbf{Implementation of CO (RQ2).} 
To demonstrate the effectiveness of each implementation detail in cache optimization, we design the following three experiments. We compare the impacts of different cache optimization positions, different cache embedding methods, and different cache hyperparameters on the generation quality.

Previous caching methods focused on Attention and MLP layers. To explore what to optimize in cached content, we study cache-optimization application locations. Table~\ref{tab3.1} shows that when cache optimization is applied to both Attention and MLP layers simultaneously, the best generation quality can be achieved (FID: 3.870 to 3.692 for FORA, 3.539 to 3.451 for L2C). But applying cache optimization to only one of them reduces generation quality, sometimes even worse than no optimization. So, synchronously optimizing both layers is crucial for leveraging this method's advantages.

\begin{figure*}[t]
\includegraphics[width=0.85\textwidth]{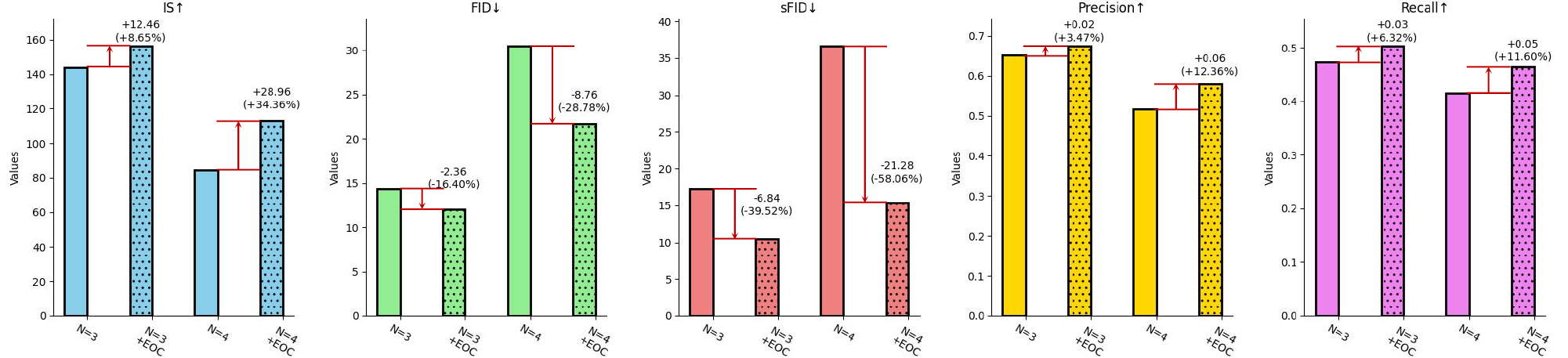}
\caption{The impact of EOC on different metrics before and after its application when $ N = 3, 4 $.} \label{fig: Ndayu2}
\end{figure*}

Regarding the selection of the mapping method for prior knowledge, by observing Table~\ref{tab3.2}, we can find that using the multiplication method in Eq.~\eqref{eq:4} to map the prior knowledge to the data distribution of the generation process can better reduce the cache error. As for methods that directly add ``trend'' to the generation process, experiments show that regardless of the value of $\theta$, the generation quality will only be worse than or infinitely close to the situation without cache optimization. Therefore, embedding cache optimization using the addition method actually introduces noise.

Finally, regarding the selection of $\theta$, we present Figure~\ref{fig: theta}. It can be found that FORA+EOC and L2C+EOC achieve the most balanced generation results when $\theta = 0.01$ and $\theta = 0.005$ respectively. For this difference, we believe that because the L2C+EOC method contains additional training information, there will be a certain deviation between the generation process and the original model, causing the $\theta$ of L2C+EOC to be smaller.

\subsection{Comparison with Other Methods}
\textbf{Effect of EOC (RQ3).} 
Next, we apply EOC to FORA and L2C with different caching intensities (25\% and 50\%) and compare the changes in generation quality before and after the application. As shown in Table~\ref{tab4.1}, compare with the original methods FORA and L2C, the methods using EOC perform better on more parameters (bolded values). Specifically, for FORA+EOC, the FID values of the models with two caching intensities decrease from 3.870 and 6.857 to 3.692 and 5.821. The sFID values decrease from 5.185 and 8.757 to 5.122 and 5.342. For L2C+EOC, the model performs better in all metrics except IS. We believe that the reason for the decrease in IS is that there is still a small amount of noise in the prior knowledge that has not been fully mapped to the features of the generation process. Meanwhile, as shown in Figure~\ref{fig: Ndayu2}, EOC has a more powerful optimization ability on generation quality when $N > 2$, and it can reduce FID by 28.78\% when $N = 4$.

\begin{figure}[t]
\includegraphics[width=0.8\linewidth]{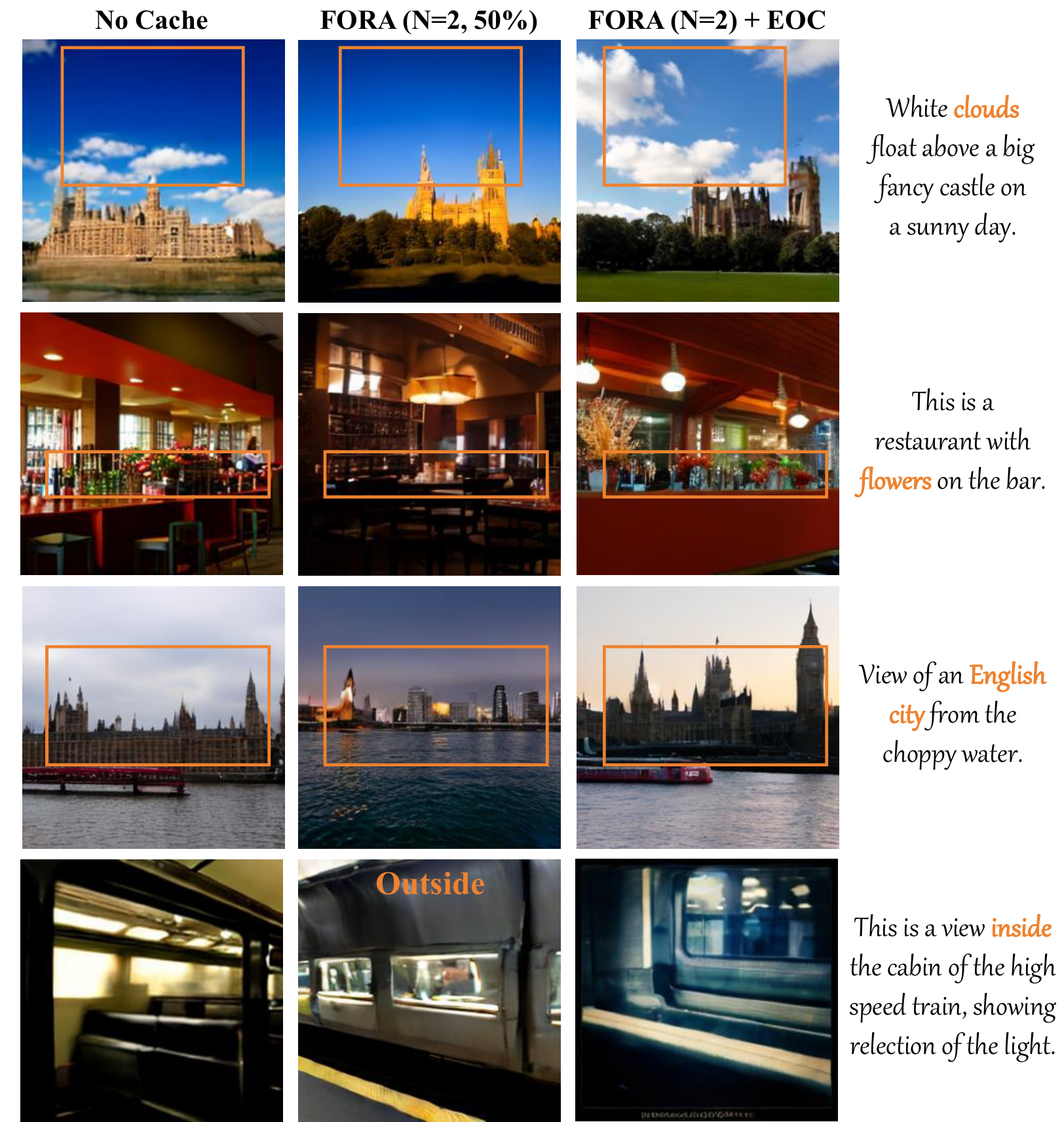}
\caption{Images generated by FORA (N=2) under Pixart-$\alpha$ before and after the application of EOC.} \label{fig: t2i}
\end{figure}

In addition, in Table~\ref{tabtoca}, EOC can improve the FID and sFID scores from 2.896 and 4.720 to 2.883 and 4.658, respectively, demonstrating the effectiveness of EOC's fine-grained caching strategy at the token level. Table~\ref{tabt2i} shows that EOC also has excellent capabilities in text-to-image tasks. Specifically, the MS-COCO FID score has been reduced from 32.637 to 32.504. In Figure~\ref{fig: t2i}, the results generated by FORA may produce hallucinations that lead to a mismatch between text and images, such as the disappearance of clouds and flowers, and the incorrect generation of "English city" and "inside the cabin" as "modern city" and "outside the cabin," respectively. EOC can fill in the missing content under caching and also resolve issues with semantic understanding errors.

Finally, we also need to compare the changes in generation time before and after the introduction of EOC. The calculation of FLOPs has already shown that under ideal conditions, the additional computational load brought by the introduction of EOC can be ignored. For the specific generation time, as shown in Table~\ref{tab4.2}, after using EOC, FORA and L2C only increased the time by about 0.56\% and 0.22\%, respectively. Compared with the 4.60\% and 2.49\% reduction in FID, this cost is reasonable and acceptable.

\begin{table}[t]
\centering
\caption{Comparison before and after incorporating EOC on the basis of DiT-ToCa-50step \cite{zou2024accelerating}.}\label{tabtoca}
\begin{tabular}{c|cc|cc}
\toprule
\textbf{Method}   & \textbf{FID↓}  & \textbf{sFID↓} & \textbf{FLOPs(T)↓} & \textbf{Speed↑} \\ \midrule
DiT \cite{peebles2023scalable} & 2.300 & 4.282 & 23.735 & 1× \\
\midrule
ToCa-50step (N=4)       & 2.896          & 4.720          &   10.233                 &      2.3194×           \\ 
ToCa-50step + EOC & \textbf{2.883} & \textbf{4.658} &       10.234             &         2.3192×        \\ \bottomrule
\end{tabular}
\end{table}

\begin{table}[t]
\centering
\caption{Comparison before and after incorporating EOC based on Pixart-FORA \cite{selvaraju2024fora}.}\label{tabt2i}
\begin{tabular}{c|c|cc}
\toprule
\textbf{Method}   & \textbf{\makecell{MS-COCO\\FID↓}}  &  \textbf{FLOPs(T)↓} & \textbf{Speed↑} \\ \midrule
Pixart-$\alpha$ \cite{chen2023pixart} & 25.959 & 11.186 & 1×
\\ \midrule
FORA-20step       & 32.637                    & 5.593 & 2×                 \\ 
FORA-20step + EOC & \textbf{32.504}  & 5.594 & 1.9996×                \\ \bottomrule
\end{tabular}
\end{table}

\begin{table}[t]
\centering
\caption{Compare the changes in the computational load for generating eight images before and after cache optimization.}\label{tab4.2}
\begin{tabular}{c|c|c}
\toprule
\textbf{Method} & \textbf{FID↓} & \textbf{latency(s)↓} \\ \midrule
FORA \cite{selvaraju2024fora} & 3.870 & 2.664 ± 0.023 \\ 
FORA + EOC & \textbf{3.692} & 2.679 ± 0.056 \\ \midrule
L2C \cite{ma2024learning} & 3.539 & 2.707 ± 0.026 \\ 
L2C + EOC & \textbf{3.451} & 2.713 ± 0.013 \\ \bottomrule
\end{tabular}
\end{table}

\section{Conclusion}
In this paper, we propose EOC, an Error-Optimized Cache method for the Diffusion Transformer (DiT). Our work extracts prior information from the DiT generation process and uses it to determine whether the cached blocks need cache optimization. Finally, we embed the variation of the prior information into the caching process, effectively reducing caching errors and preventing error accumulation. The proposed EOC can generate samples of higher quality with little increase in computational cost. Through model comparison and ablation experiments, we achieve better performance under FORA and L2C with different caching intensities, which also verifies the effectiveness and necessity of the innovations we proposed. 

We find that the introduction of EOC has a slight negative impact on some parameters (such as IS) under certain caching strategies. Therefore, we plan to further investigate noise filtering and the embedding methods for cache optimization in the future.



\section*{Acknowledgements}
This research is supported by the Local Science and Technology Program (No.2024CSJGG00800).

\balance
\bibliographystyle{ACM-Reference-Format}
\bibliography{ref}




\end{document}